\documentclass[letterpaper, 10 pt, conference]{ieeeconf}

\IEEEoverridecommandlockouts                              

\usepackage{myStyle}

\begin{document}

\title{\LARGE \bf
Biased-MPPI: Informing Sampling-Based Model Predictive Control\\ by Fusing Ancillary Controllers
}

\author{Elia Trevisan and Javier Alonso-Mora
\thanks{This research is supported by the project ``Sustainable Transportation and Logistics over Water: Electrification, Automation and Optimization (TRiLOGy)'' of the Netherlands Organization for Scientific Research (NWO), domain Science (ENW), and the Amsterdam Institute for Advanced Metropolitan Solutions (AMS) in the Netherlands.}
\thanks{The authors are with the Cognitive Robotics Department,
        TU Delft, 2628 CD Delft, The Netherlands
        {\tt\small \{e.trevisan, j.alonsomora\}@tudelft.nl}}
}

\maketitle
\thispagestyle{empty}
\pagestyle{empty}

\begin{abstract}

Motion planning for autonomous robots in dynamic environments poses numerous challenges due to uncertainties in the robot's dynamics and interaction with other agents. Sampling-based MPC approaches, such as Model Predictive Path Integral (MPPI) control, have shown promise in addressing these complex motion planning problems. However, the performance of MPPI relies heavily on the choice of sampling distribution. Existing literature often uses the previously computed input sequence as the mean of a Gaussian distribution for sampling, leading to potential failures and local minima. In this paper, we propose a novel derivation of MPPI that allows for arbitrary sampling distributions to enhance efficiency, robustness, and convergence while alleviating the problem of local minima. We present an efficient importance sampling scheme that combines classical and learning-based ancillary controllers simultaneously, resulting in more informative sampling and control fusion. Several simulated and real-world demonstrate the validity of our approach.
\flushleft{Website: \href{https://autonomousrobots.nl/paper_websites/biased-mppi}{autonomousrobots.nl/paper\_websites/biased-mppi}}
\end{abstract}


\section{Introduction}\label{sec::intro}
Navigating autonomous robots through dense and dynamic environments poses a formidable challenge due to significant uncertainties, including the robot's state, model, environmental conditions, and interactions with other agents. Achieving desired behaviors under such conditions often necessitates using intricate cost functions and constraints, resulting in complex, nonlinear, non-convex, and occasionally discontinuous problem formulations. The dynamic nature of the environment introduces potential unexpected changes, demanding rapid adaptability in the robot's actions.

To address these challenges, one approach is to cast the problem in a stochastic optimal control setting, where they can be mathematically represented as stochastic Hamilton-Jacobi-Bellman (HJB) equations. However, solving these equations numerically can be challenging due to the curse of dimensionality.
Pioneering work demonstrated that the stochastic HJB equations can be linearized for control-affine systems, and their solution can be approximated through sampling using the path integral formulation~\cite{kappen_path_2005}.
Implemented in a receding horizon fashion, Model Predictive Path Integral (MPPI) control~\cite{williams_aggressive_2016, williams_model_2017}, and its Information-Theoretic counterpart~\cite{williams_information_2017, williams_information-theoretic_2018} have been initially used for racing a small-scale rally car. MPPI has also been successfully applied to several other planning problems, such as for autonomous vehicles with dynamic obstacles~\cite{perez-morales_Information-Theoretic_2021}, solving games~\cite{williams_best_2018}, flying drones in partially observable environments~\cite{mohamed_model_2020}, performing complex maneuvers~\cite{pravitra_flying_2021} and used in combination with adaptive control schemes~\cite{pravitra_l1-adaptive_2020}. It has also been adapted to multi-agent systems for formation flying~ \cite{gomez_real-time_2016}, cooperative behavior~\cite{wan_cooperative_2021}, and simultaneous prediction and planning~\cite{streichenberg_multi-agent_2023}. Furthermore, MPPI has shown promise in manipulating objects with robot arms\cite{bhardwaj_storm_2021} including model uncertainties ~\cite{abraham_model-based_2020}, in pushing tasks~\cite{arruda_uncertainty_2017, cong_self-adapting_2020} and planning motion for four-legged walking robots~\cite{carius_constrained_2021}.
MPPI is a model-based approach that requires a model to forward simulate trajectories given sampled inputs. Recent work has utilized physics engines to simulate samples~\cite{howell_predictive_2022, pezzato_sampling-based_2023-1}, eliminating the need for explicitly defining the dynamics of agents and the environment, thus providing a significant advantage in contact-rich manipulation tasks.

\begin{figure}[!t]
    \centering
    \includegraphics[width=0.48\textwidth]{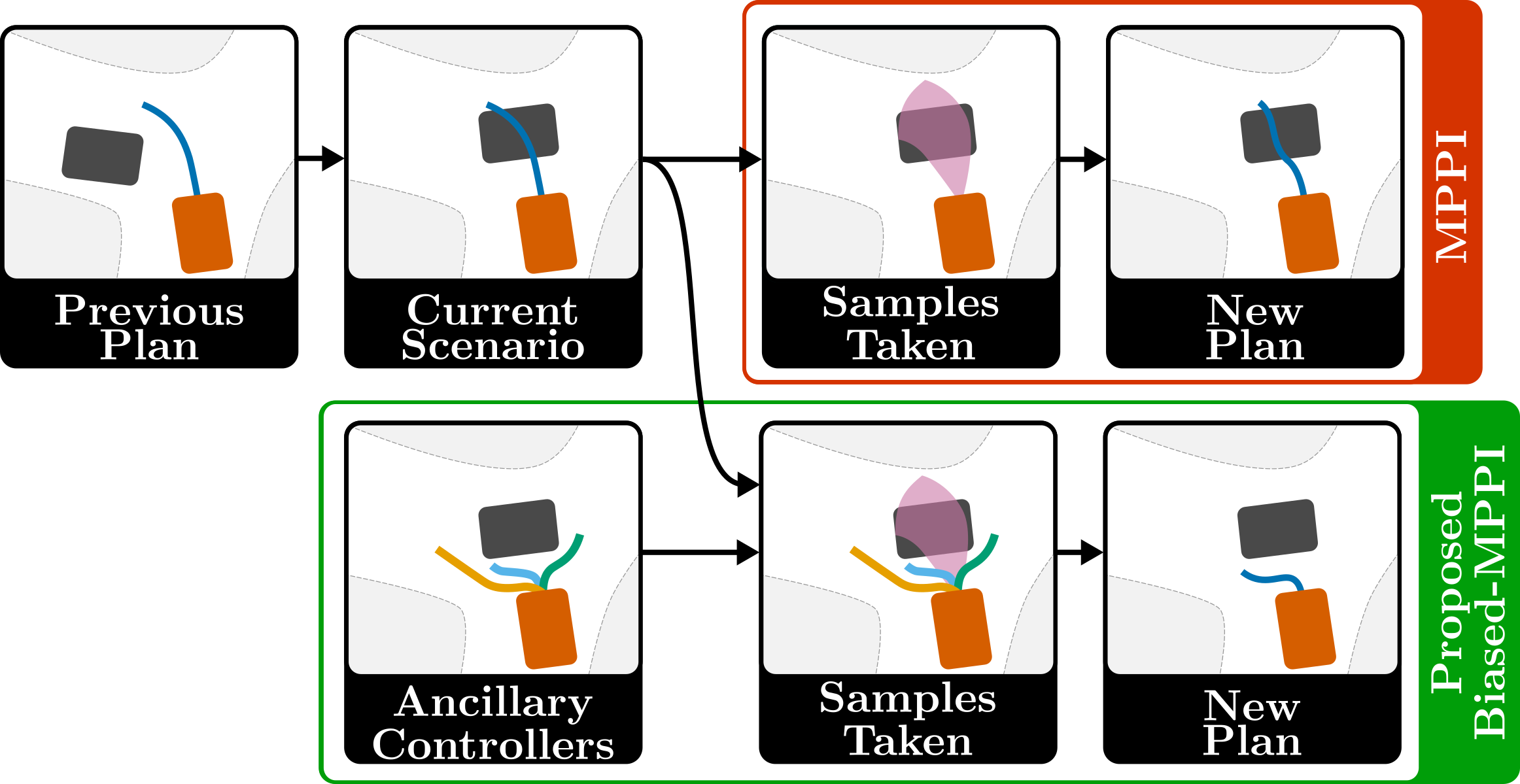}
    \caption{\textit{Top}: Usually, MPPI only takes samples around a previous plan. Here, the environment changes unexpectedly, and all the sampled trajectories are in collision, which leads to computing a new plan that also collides. \textit{Bottom}: our Biased-MPPI adds ancillary controllers to the sampling distribution, quickly converging to a collision avoidance maneuver.}
    \label{fig:sampling_fail}
\end{figure}

One of the critical challenges in applying MPPI to dynamic environments is ensuring the algorithm's performance and reliability. The success of MPPI heavily relies on the choice of sampling distribution, which is crucial, especially in real-time scenarios. Most existing literature uses the previously computed input sequence as the mean of a Gaussian distribution for sampling~\cite{williams_aggressive_2016}, with the variance being tunable. However, using the previous input sequence may trap the algorithm in local minima. This can lead to catastrophic failures in the presence of unexpected disturbances or changes in the environment~\cite{williams_robust_2018} (\Cref{fig:sampling_fail}).

This paper explores the application of MPPI in dynamic environments, emphasizing the need to improve its performance and reliability in the face of unexpected disturbances and rapidly changing conditions.

\subsection{Previous Work}\label{sec::prev_work}
Several works tried to make the method more efficient or more robust. Early work \cite{mensink_ep_2010} proposed using Expectation Propagation instead of Monte Carlo sampling, demonstrating better efficiency in scenarios with hard constraints. Other works instead accelerate the convergence of MPPI by leveraging gradient descent updates \cite{okada_acceleration_2018}. Another option to be more reactive to environmental changes is to iteratively converge to a solution through adaptive importance sampling \cite{asmar_model_2023}. This, however, requires multiple iterations between each planning time step, diminishing the parallelizability of MPPI.
Many other works propose improving the algorithm's convergence by somehow changing its sampling distribution. This can be done by substituting the Gaussian used for sampling with a different hand-crafted distribution \cite{mohamed_autonomous_2022} or by directly learning a distribution from data \cite{kusumoto_informed_2019, power_variational_2022}.
Given that MPPI allows for tuning the variance of the sampling distribution~\cite{williams_model_2017}, some works sought to improve the efficiency of the scheme by adapting the covariance online via covariance steering~\cite{balci_constrained_2022, yin_trajectory_2022}.
Other ways to improve efficiency can be to fit splines to the sampled inputs~\cite{bhardwaj_storm_2021} or to constrain the distribution to sample areas that are known to contain low-cost trajectories~\cite{carius_constrained_2021}.
Previous works have also experimented with ancillary controllers. In~\cite{arslan_information-theoretic_2014}, authors propose to sample inputs around a path previously computed by RRT. Other works instead robustify MPPI by switching to an iLQG controller~\cite{williams_robust_2018} or by integrating one into the system's model~\cite{gandhi_robust_2021}.
Previous work also compares an MPPI that samples around a previously computed input, an input sequence computed by a sequential linear-quadratic MPC, and a learned sampling policy~\cite{carius_constrained_2021}.
In general, however, the original derivations of MPPI~\cite{williams_information-theoretic_2018} only allow samples to be drawn from a uni-modal Gaussian distribution, usually centered around the previous control sequence, which can hamper performance and reduce reactivity to unexpected changes in the environment.

\subsection{Contributions}

We propose a Biased-MPPI, for which we provide mathematical derivations that allow for arbitrary changes to the sampling distribution. We discuss the impact of introducing biases in the sampling distribution on the overall method. We experiment with an importance sampler that utilizes multiple classical and learning-based ancillary controllers simultaneously to take more informative samples, which can be seen as a control fusion scheme. Through simulated and real-world experiments, we demonstrate the impact of taking suggestions from several underlying controllers on robustness to model uncertainties and local minima, reactivity to unexpected events, and sampling efficiency.


\section{Preliminaries}
In this section, we provide a concise introduction to the key concepts of MPPI within the Information-Theoretic framework. For more details, we direct the reader to prior research~\cite{williams_information-theoretic_2018}. We begin by defining a function:
\begin{equation}
    \mathcal{F}(S, \mathbb{P}, x_0, \lambda)  = -\lambda \log \left(\mathbb{E_P}\left[ \text{exp}\left(-\frac{1}{\lambda} S(V)\right)\right] \right)
\end{equation}
which we will denote as the free energy of the system. Here, $V$ represents a sequence of inputs, $\mathbb{P}$ is a base measure, $\lambda$ is a tuning parameter, $S(V)$ is a cost, and $x_0$ represents the system's initial state. It can be shown that: 
\begin{equation} \label{eq::free_energy_ineq}
    \mathcal{F}(S, \mathbb{P}, x_0, \lambda) \leq \mathbb{E_Q}[S(V)] + \lambda \mathbb{KL}(\mathbb{Q}||\mathbb{P}).
\end{equation}
Here, $\mathbb{Q}$ represents a probability measure that characterizes the controlled input distribution, and $\mathbb{KL}(\mathbb{Q}||\mathbb{P})$ denotes the KL-Divergence between the base measure and the controlled measure.~\cref{eq::free_energy_ineq} signifies that the free energy serves as a lower bound for the expected cost under the controlled distribution plus a control cost represented by the KL-Divergence.
Hence, determining a control distribution that achieves this lower bound minimizes the expected cost and control cost.
We can define a control distribution $\mathbb{Q}^*$ through its Radon-Nikodym derivative to the base measure:
\begin{equation}\label{eq::RN_der}
    \frac{d\mathbb{Q^*}}{d\mathbb{P}} = \frac{\exp(-\frac{1}{\lambda} S(V))}{\mathbb{E_P}[\exp(-\frac{1}{\lambda} S(V))]}.
\end{equation}
Substituting $\mathbb{Q}$ with $\mathbb{Q}^*$ in \cref{eq::free_energy_ineq}, we can prove that $\mathbb{Q}^*$ is an optimal control distribution in the sense that it achieves the lower bound.
The idea is now to align our control distribution $\mathbb{Q}$ with the optimal distribution $\mathbb{Q}^*$ though KL minimization, which results in the optimal input sequence $U^*$:
\begin{equation}\label{eq::info_optimal_input}
    U^* = \argmin_U \mathbb{KL(Q^*||Q)}.
\end{equation}
Now, considering a discrete-time system:
\begin{equation}
    x_{t+1} = F(x_t, v_t),   \quad   v_t \sim \mathcal{N}(u_t,\Sigma).
\end{equation}\label{sec::preliminaries}
Here, $x_t \in \mathbb{R}^n$ represents the state vector at time step $t$, $F(\cdot)$ is the state transition model, $v_t \in \mathbb{R}^m$ denotes the noisy input, $u_t \in \mathbb{R}^m$ is the commanded input, and $\Sigma$ corresponds to the natural input variance of the system. If $\mathbb{P}$ and $\mathbb{Q}$ are the uncontrolled and controlled measures, respectively, we can define them through their probability density functions:
\begin{equation*}
    \begin{aligned}
    p(V) &= \prod_{t=0}^{T-1} \frac{1}{((2\pi)^m |\Sigma|)^{1/2}} \exp \left( -\frac{1}{2} v_t^T \Sigma^{-1} v_t \right)\\
    \medmath{q(V|U)} &\medmath{= \prod_{t=0}^{T-1} \frac{1}{((2\pi)^m |\Sigma|)^{1/2}} \exp \left( -\frac{1}{2} (v_t-u_t)^T \Sigma^{-1} (v_t-u_t) \right)}.
    \end{aligned}
\end{equation*}
It can be proven from~\cref{eq::info_optimal_input} that the optimal control input at time $t$ is the mean input under the optimal distribution:
\begin{equation}\label{eq::Williams_optimal_in}
    u_t^* = \int_{\Omega_V} q^*(V) v_t dV.
\end{equation}
We can estimate such mean sampling from our controlled distribution via importance sampling:
\begin{equation}\label{eq::estimate_optimal_input}
\begin{aligned}
        u_t^* &= \int \frac{q^*(V)}{q(V|U)} q(V|U) v_t dV\\
        &= \mathbb{E_Q}[\omega(V) v_t],
\end{aligned}
\end{equation}  
with the importance sampling weight $\omega(V)$ being:
\begin{equation}
\begin{aligned}\label{eq::original_weight}
    &\omega(V) = \left( \frac{q^*(V)}{q(V|U)} \right) = \left( \frac{q^*(V)}{p(V)} \right) \left(\frac{p(V)}{q(V|U)}\right)\\ 
    &\medmath{=\frac{1}{\eta} \exp \left( - \frac{1}{\lambda} \left( S(V) + \frac{\lambda}{2} \sum_{t=0}^{T-1} u_t^T \Sigma^{-1} u_t + 2u_t^T \Sigma^{-1} \epsilon_t \right) \right)}.\\
\end{aligned}
\end{equation}
We can, therefore, sample $K$ noisy input sequences:
\begin{equation}\label{eq::classic_mppi_samples}
    \begin{aligned}
        V^k &= [v_0^k, v_1^k,\cdots,v_t^k, \cdots, v_{T_H}^k]\\
        v_t^k & \sim \mathcal{N}(u_{t},\Sigma)
    \end{aligned}
\end{equation}
where $t$ is a time step and $T_H$ is the planning horizon. A practical choice often made in MPPI is to take $u_t$ as a time-shifted version of the previously computed approximation of the optimal control sequence.
We roll out the sampled $V^k$ into state trajectories using the system's model $F(\cdot)$, evaluate their cost $S(V)$, compute the weights $\omega(V)$, get a new estimate of the optimal input sequence $U^*$ via~\cref{eq::estimate_optimal_input} and iterate.
In~\cref{eq::original_weight}, the control cost is multiplied and divided by $\lambda$. Not having control over the magnitude of the terms at the exponential can cause numerical issues. A change of base measure $\mathbb{P}$ can solve the problem~\cite{williams_information-theoretic_2018}. One might also need a higher variance $\Sigma_s$ for sampling compared to the natural variance of the system $\Sigma$~\cite{williams_model_2019}. This again introduces terms at the exponential independent from $\lambda$. Moreover, introducing an arbitrary, potentially multi-modal sampling distribution $\mathbb{Q}_s$ is difficult. All these issues stem from the ratio $p(v)/q(V|U)$ in \cref{eq::original_weight}. Our approach addresses this by showing that accepting a bias in the solution can eliminate the ratio and allow for arbitrary sampling distributions.


\section{Proposed Approach}\label{sec::prop_approach}

\subsection{Biased-MPPI}
Let us first redefine the cost function as:
\begin{equation}\label{eq::s_tilde}
    \Tilde{S}(V) = S(V) + \lambda \log \left( \frac{p(V)}{q_s(V)}\right).
\end{equation}
We then define the free-energy with this new cost:
\begin{equation}
\begin{aligned}
    &\mathcal{F}(\Tilde{S}, \mathbb{P}, x_0, \lambda)  = -\lambda \log \left(\mathbb{E_P}\left[ \text{exp}\left(-\frac{1}{\lambda} \Tilde{S}(V)\right)\right] \right)\\
    &=-\lambda \log \left(\mathbb{E_Q}\left[ \text{exp}\left(-\frac{1}{\lambda} \Tilde{S}(V)\right) \frac{p(V)}{q(V)}\right] \right)\\
    &\leq -\lambda \mathbb{E_Q}\left[  \log \left( \text{exp}\left(-\frac{1}{\lambda} \Tilde{S}(V)\right) \frac{p(V)}{q(V)} \right) \right] = *
\end{aligned}
\end{equation}
where, as in~\cite{williams_information-theoretic_2018}, we applied Jensen's inequality. We can simplify the right-hand side as follows:
\begin{equation*}
\begin{aligned}
    &* =-\lambda \mathbb{E_Q}\left[ -\frac{1}{\lambda} \Tilde{S}(V) + \log \left(\frac{p(V)}{q(V)} \right) \right]\\
    &=-\lambda \mathbb{E_Q}\left[ -\frac{1}{\lambda} S(V) - \log \left(\frac{p(V)}{q_s(V)} \right) + \log \left(\frac{p(V)}{q(V)} \right) \right]\\
    &= \mathbb{E_Q}\left[S(V)\right] + \lambda \mathbb{E_Q}\left[ \log \left(\frac{p(V)}{q_s(V)} \frac{q(V)}{p(V)}\right) \right]\\
    &= \mathbb{E_Q}\left[S(V)\right] + \lambda \mathbb{KL}(\mathbb{Q}||\mathbb{Q}_s).
\end{aligned}
\end{equation*}
The free energy inequality is then:
\begin{equation}\label{eq::bias}
    \mathcal{F}(S, \mathbb{P}, x_0, \lambda) \leq \mathbb{E_Q}\left[S(V)\right] + \lambda \mathbb{KL}(\mathbb{Q}||\mathbb{Q}_s).
\end{equation}
Thus, while we start with $\Tilde{S}(V)$, the free energy serves as a lower bound for the expected original cost $S(V)$ under the controlled distribution plus lambda times the KL-Divergence between the controlled and sampling distribution. An optimal control distribution achieving the lower bound would minimize the original cost $S(V)$ while pushing the controlled distribution to align with the sampling distribution, effectively introducing a bias toward the sampling distribution. We define a controlled distribution $\mathbb{Q}^*$ as:
\begin{equation*}
    \frac{d\mathbb{Q^*}}{d\mathbb{P}} = \frac{\exp(-\frac{1}{\lambda} \Tilde{S}(V))}{\mathbb{E_P}[\exp(-\frac{1}{\lambda} \Tilde{S}(V))]}.
\end{equation*}
Under $\mathbb{Q}^*$, the KL-Divergence becomes:
\begin{equation*}
\begin{aligned}
    &\mathbb{KL}(\mathbb{Q}^*||\mathbb{Q}_s) = \mathbb{E_{Q^*}}\left[ \log \left(\frac{q^*(V)}{q_s(V)} \right) \right]\\
    & = \mathbb{E_{Q^*}}\left[ \log \left(\frac{q^*(V)}{p(V)} \right) \right] + \mathbb{E_{Q^*}}\left[ \log \left(\frac{p(V)}{q_s(V)} \right) \right]\\
    & = -\frac{1}{\lambda} \mathbb{E_{Q^*}}\left[ \Tilde{S}(V) \right] 
    - \log \left(\mathbb{E_P}\left[ \text{exp}\left(-\frac{1}{\lambda} \Tilde{S}(V)\right)\right] \right)\\
    & \quad \quad + \mathbb{E_{Q^*}}\left[ \log \left(\frac{p(V)}{q_s(V)} \right) \right]
\end{aligned}
\end{equation*}
Substituting into~\cref{eq::bias} and simplifying leads to:
\begin{equation*}
\begin{aligned}
    \mathcal{F}(\Tilde{S}, \mathbb{P}, x_0, \lambda)  & \leq -\lambda \log \left(\mathbb{E_P}\left[ \text{exp}\left(-\frac{1}{\lambda} \Tilde{S}(V)\right)\right] \right)\\
    & = \mathcal{F}(\Tilde{S}, \mathbb{P}, x_0, \lambda).
\end{aligned}
\end{equation*}
This proves that $\mathbb{Q}^*$ is the optimal distribution in that it achieves the lower bound in~\cref{eq::bias}. Following the steps in~\cite{williams_information-theoretic_2018}, we can align our controlled distribution $\mathbb{Q}$ to $\mathbb{Q}^*$ as in~\cref{eq::Williams_optimal_in}, except we can now use our sampling distribution:
\begin{equation}
u_t^* = \mathbb{E}_{\mathbb{Q}_s}[\omega(V) v_t],
\end{equation}
with importance sampling weights:
\begin{equation}
\begin{aligned}
    &\omega(V) =\frac{1}{\eta} \exp \left( - \frac{1}{\lambda} \Tilde{S}(V)\right) \left(\frac{p(V)}{q_s(V)}\right)\\
    &=\frac{1}{\eta} \exp \left( - \frac{1}{\lambda} \left(S(V) + \lambda \log \left( \frac{p(V)}{q_s(V)}\right)\right)\right) \\
    & \quad \quad \quad \quad \times \exp \left( \log \left(\frac{p(V)}{q_s(V)}\right)\right)\\
    &= \frac{1}{\eta} \exp \left( - \frac{1}{\lambda} S(V)\right).
\end{aligned}
\end{equation}
Note that our change of cost~\cref{eq::s_tilde} resulted in the optimal control being biased towards the sampling distribution, as shown in~\cref{eq::bias}. However, this simplified the weights $\omega(V)$ and allowed us to design arbitrary sampling distributions $\mathbb{Q}_s$.
In~\cite{williams_information-theoretic_2018}, $S(V)$ was defined as the state-dependent cost. However, this restriction was made to relate the approach to path integral control~\cite{kappen_path_2005}. Such relation was only shown exactly when $\mathbb{P}$ is the distribution induced by an uncontrolled continuous-time control-affine system. This restriction is not required in the Information-Theoretic framework, which allows for a larger class of systems, and one can add input costs in S(V).

\subsection{Sampling from Ancillary Controllers}
There are several ways one could design an arbitrary sampling distribution. This paper focuses on taking most samples around a previously computed input distribution and some samples from hand-crafted policies.

In particular, we design a set of task-specific ancillary controllers, these being, e.g., open-loop motion primitives, reference tracking feedback controllers, or learning-based strategies to propose $J$ input sequences $U^j = [u_{0}^j, u_{1}^j,\dots,u_{t}^j, \dots, u_{T_H}^j]$. These ancillary controllers are described for each experiment in~\Cref{sec::ill_ex,sec::experiments}. We then choose the K sampled input sequences $V_s^k$ as,

\begin{equation}
V_s^k= \left\{
\begin{aligned}
      & U^j, \text{ with } j=k  \quad &\text{if } k\leq J\\
      & V^k, \text{ as in~\cref{eq::classic_mppi_samples}} & \text{if } k>J,\\
\end{aligned} 
\right.
\end{equation}
meaning that, at each time step, we take one sample from each of the $J$ ancillary controllers, and the remaining $K-J$ samples are taken according to the classical MPPI strategy.

\subsection{Autotuning the Inverse Temperature}
As in~\cite{pezzato_sampling-based_2023-1} and similarly to~\cite{carius_constrained_2021}, we autotune the inverse temperature $\lambda$ online based on the normalization factor $\eta$.
\begin{equation}
    \label{eq:lambda_update}
    \lambda_{t+1} = 
    \begin{cases}
      0.9\lambda_t    & \text{if}\  \eta > \eta_{max} \\
      1.2\lambda_t      & \text{if}\  \eta < \eta_{min}\\
      \lambda_t      & \text{otherwise}
    \end{cases}
\end{equation}
In all experiments, this can roughly keep the number of samples with a significant weight between $\eta_{min}$ and $\eta_{max}$.

\section{Illustrative Experiment}\label{sec::ill_ex}
\begin{figure}[h]
    \centering
    \includegraphics[width=0.4\textwidth]{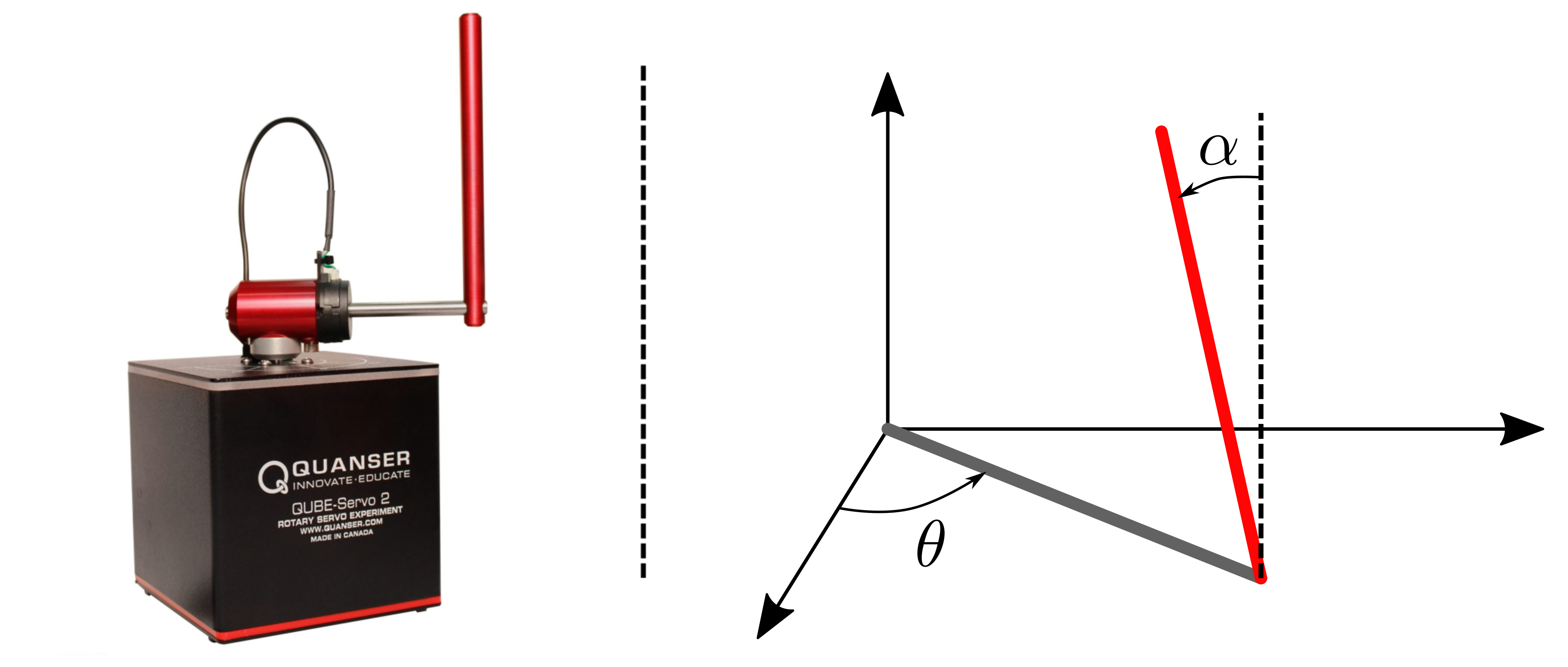}
    \caption{\textit{Left}, Quanser Qube-Servo, and \textit{right}, its diagram. The arm's rotation, $\theta$, is the actuated angle. The angle between the pendulum and the upright position, $\alpha$, is not actuated.}
    \label{fig:qube_servo}
\end{figure}

We apply our Biased-MPPI to a rotary inverted pendulum \cite{quanser_inc_qube_nodate} (\Cref{fig:qube_servo}) in simulation to visualize its main features.

\begin{figure*}[!h]
    \medskip
  \centering
  \includegraphics[width=\textwidth]{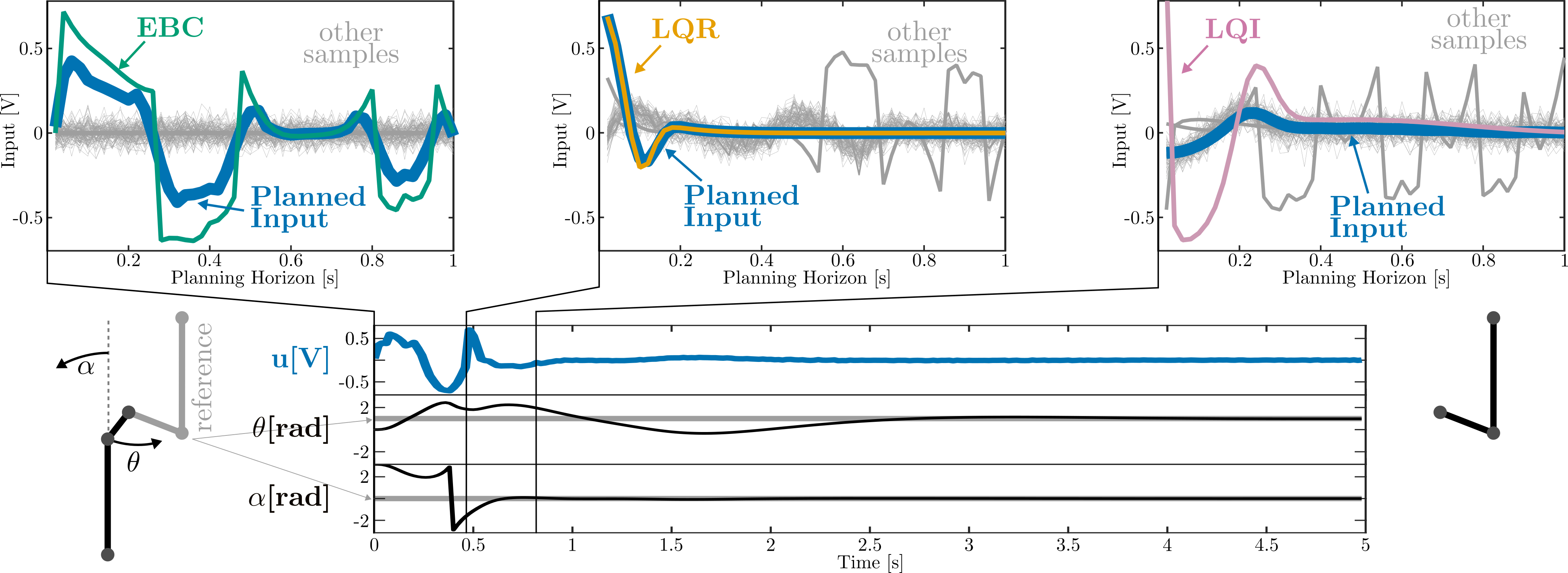}
  \caption{Input and state evolution during a pendulum experiment with Biased-MPPI. We show the samples taken and the resulting planned input sequence over the planning horizon for three instances. While we sample all ancillary controllers in each instance, we highlight the one with the most influence on the planned input sequence.}
  \label{fig:pendulum_inputs}
\end{figure*}

\subsection{Swing-up and tracking}
Starting at the bottom equilibrium with $\theta_0 = 0$ and $\alpha_0 = \pi$, the task is to swing up the pendulum to $\alpha_r = 0$ while keeping the arm close to $\theta_r = 1$. Thus, the running cost is:
\begin{equation}\label{eq:pendulum_cost}
     C_p(x(t)) = 100((\theta_t - \theta_r)^2 + (\alpha_t - \alpha_r)^2) + \dot{\theta_t}^2 +2\dot{\alpha_t}^2.
\end{equation}
The system has dynamics $x(t+1) = F(x(t), u(t))$, where the state of the system at time-step $t$ is denoted as $x(t) = [\theta_t, \alpha_t, \dot{\theta_t}, \dot{\alpha_t}]^T$, and $u$ represents the system's input. The nonlinear model is derived from the Lagrange equations. To design linear controllers, the model is linearized at the top equilibrium using Euler-Lagrange's method~\cite{tejado_physical_2016}.
To showcase resilience against model uncertainties, the parameters of the simulation's pendulum model are multiplied by $1+\gamma$ in each experiment, where $\gamma \sim \mathcal{N}(0, 0.05)$. The seed is consistent across methods.
The system is dicretized and controllers run at $50Hz$, the controller plans $T_H=50$ steps ahead ($1s$), covariance $\Sigma_s=0.5$, $\eta_{min}=2$ and $\eta_{max}=5$.

\subsubsection{Ancillary Controllers}
We design three ancillary controllers as a baseline and to guide the sampling strategy.

\paragraph{A Linear Quadratic Regulator (LQR)} designed using the \verb"lqr" command in Matlab, stabilizes the pendulum at the top equilibrium.


\paragraph{A Linear Quadratic Integral (LQI)} tracks the reference $\theta_r$ while maintaining the pendulum at the top equilibrium. It is synthesized with the \verb"lqi" command in Matlab.


\paragraph{A nonlinear Energy-Based Controller (EBC)} is designed as in~\cite{tejado_physical_2016} to swing up the pendulum to the top equilibrium by increasing the potential energy of the system~\cite{astrom_swinging_2000}.


\subsubsection{Switching Controller}
We introduce as baseline a switching strategy \cref{eq::switching strategy} that combines all ancillary controllers. It swings up the pendulum using the input from the ECB, $u_{ebc}$, until $\alpha$ is within $\alpha_{catch}=0.2$ of the top equilibrium. The LQR controller, with $u_{lqr}$, then stabilizes the pendulum. Once the pendulum is close to the top equilibrium ($\alpha_{track}=0.05$) with angular velocity below $\dot{\alpha}_{track}=0.1$ rad/s, the LQI, with $u_{lqi}$, is engaged for reference tracking.
\begin{equation}\label{eq::switching strategy}
u= \left\{
\begin{aligned}
      & u_{lqi}, \quad \text{if } (|\alpha|<\alpha_{track})\cap(|\dot{\alpha}|<\dot{\alpha}_{track})\\
      & u_{lqr}, \quad \text{if } (|\alpha|<\alpha_{catch})\\
      & u_{ebc}, \quad \text{otherwise} \\
\end{aligned} 
\right. 
\end{equation}

\subsubsection{Results}


\Cref{fig:pendulum_inputs} depicts a pendulum experiment's input and state evolution with Biased-MPPI, also showcasing the samples taken and the ancillary controllers' influence on the plan. At the beginning of the experiment, ECB rapidly swings up the pendulum, heavily influencing Biased-MPPI's planned input. Once near equilibrium, LQR provides a stabilizing sequence, closely tracked by Biased-MPPI. As stability is achieved, LQI suggests an input sequence swiftly bringing the arm towards the reference, albeit with high velocities. Hence, Biased-MPPI, while influenced by LQI, opts for a lower amplitude input sequence due to cost function~\cref{eq:pendulum_cost}.

\begin{figure}
    \centering
    \includegraphics{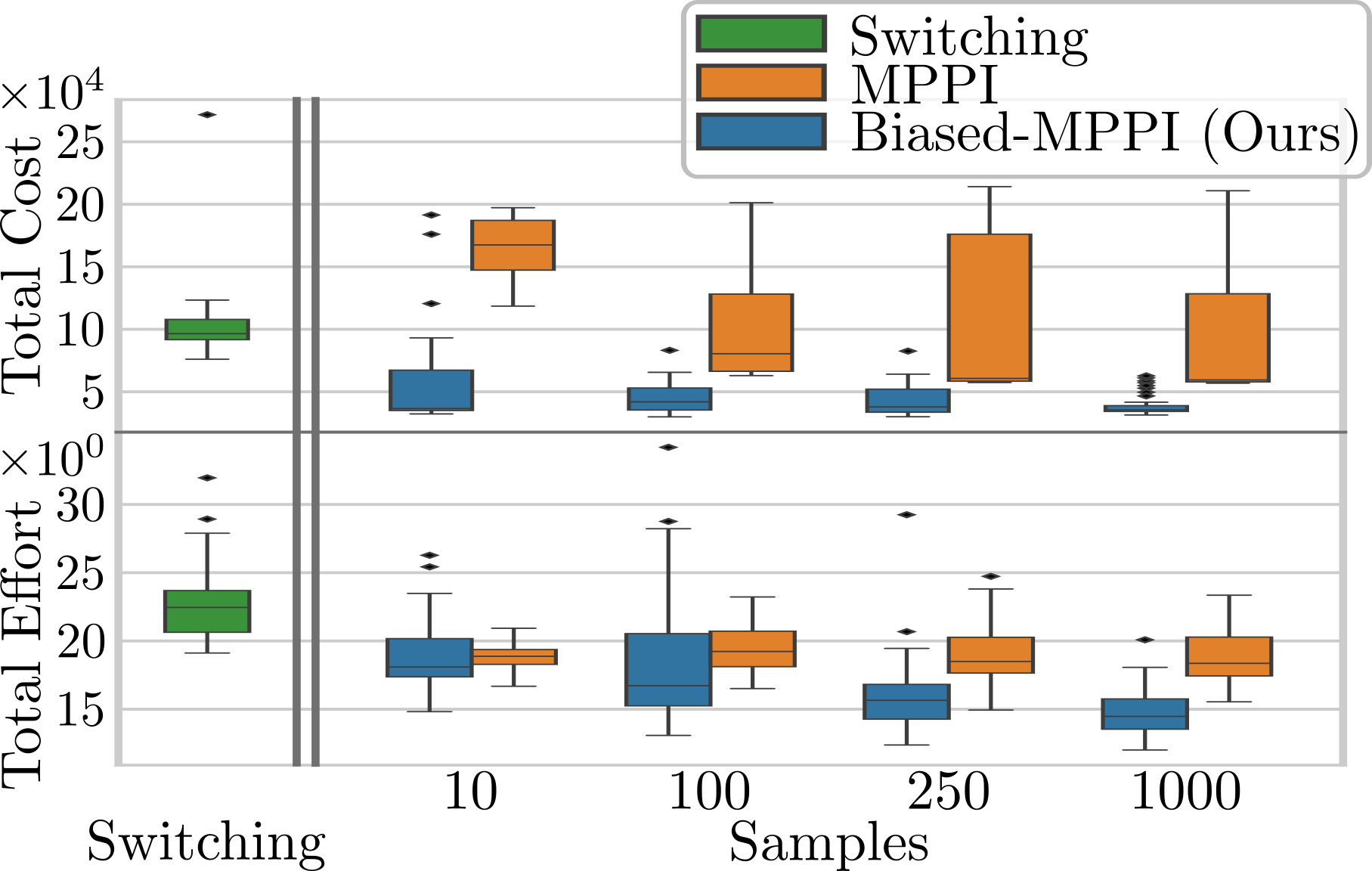}
    \caption{Total cost and control effort over 50 pendulum swing-ups with randomized model parameters.}
    \label{fig:pendulum_boxplot}
\end{figure}

\Cref{fig:pendulum_boxplot} displays the distribution of total costs, defined as $\sum_{t=0}^{T_{end}}C_p(x(t))$ where $T_{end}=250$ ($5s$) is the end of the episode, and the distribution of total efforts, defined as $\sum_{t=0}^{T_{end}}|u(t)|$, across 50 experiments. Biased-MPPI consistently outperforms both the switching strategy and the classic MPPI, regardless of the number of samples used. Moreover, the results indicate that including ancillary controllers in the proposed Biased-MPPI vastly improves the sampling efficiency, requiring fewer samples for better performance and enhancing the algorithm's robustness to model uncertainties.

\section{Simulated Motion Planning Experiments}\label{sec::experiments}
Interaction-Aware (IA) MPPI~\cite{streichenberg_multi-agent_2023} is a decentralized communication-free motion planning method that predicts short-term goals of other agents with a constant velocity model and, under homogeneity and rationality assumptions, each agent simultaneously plans and predicts motions for all agents.
In its cost function, IA-MPPI encourages adherence to navigation rules, such as giving the right-of-way to agents from the right and preferring the right-hand side during head-on encounters.
We will investigate the effects of biasing its sampling scheme with ancillary controllers.
The agents are vessels modeled using Roboat's model~\cite{wang_design_2018}. Controllers run at $10Hz$, plan $T_H=100$ steps ahead ($10s$), with $\Sigma_s=diag(6,\text{ }6,\text{ }0.12,\text{ }0.12)$, $\eta_{min}=5$ and $\eta_{max}=10$.

\subsection{Solving an Intersection}
\begin{figure*}[ht!]
    \medskip
\centering
\begin{subfigure}[c]{\textwidth}
   \includegraphics[width=1\linewidth]{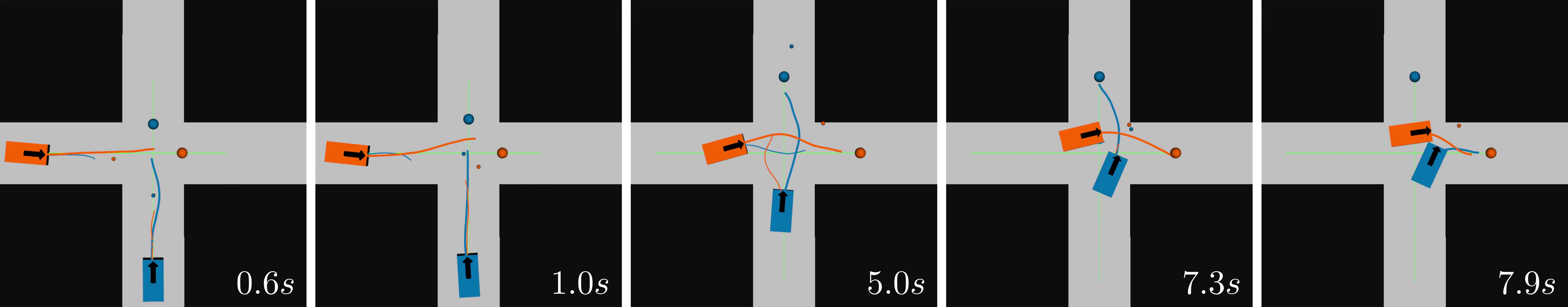}
   \caption{Using a classical MPPI sampling scheme, the agents remain in a local minimum where both want to pass first, resulting in a collision.}
   \label{fig:crossing_vanilla} 
\end{subfigure}

\begin{subfigure}[c]{\textwidth}
   \includegraphics[width=1\linewidth]{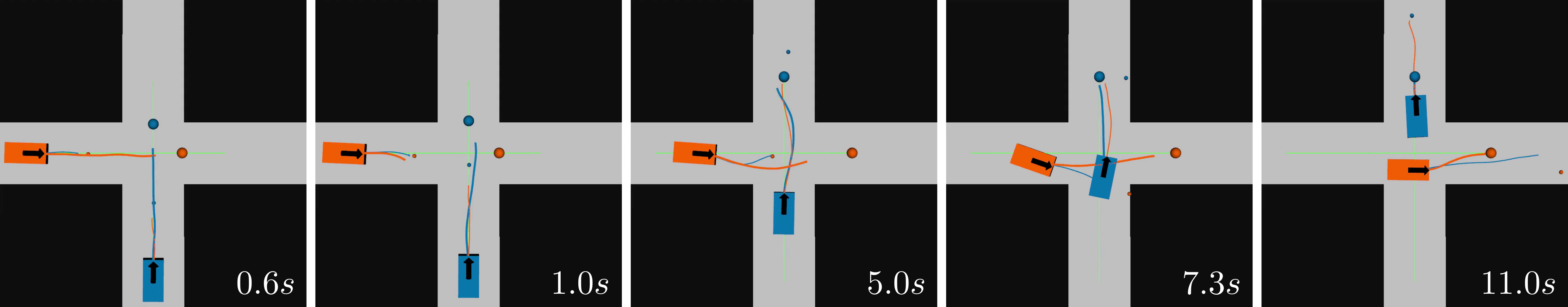}
   \caption{Using the proposed Biased-IA-MPPI, the orange agent gives way to the blue agent as soon as it is clear that both agents want to cross.}
   \label{fig:crossing_biased}
\end{subfigure}

\caption{Two vessels cross each other's path while penalized when not giving the right-of-way to agents coming from their right. The large circles are the agents' true local goals extracted from a global path. IA-MPPI is decentralized and communication-free, so the small dots are the goals vessels estimate of one another using constant velocity. The trajectories in blue are those the blue agent has planned for itself and predicted for the other, and the same goes for the orange agent.}
   \label{fig:crossing} 
\end{figure*}
An issue that can arise with classical MPPI formulation, which only takes samples around what was previously considered to be optimal, is the difficulty, once in one, of jumping out of local minima. This is particularly evident in IA-MPPI, especially in a crossing scenario. 
In this experiment, depicted in~\Cref{fig:crossing}, two identical vessels start with zero velocity and have to cross each other's paths. In their cost function, described in previous work~\cite{streichenberg_multi-agent_2023}, the decentralized and communication-free IA-MPPI is encouraged to get each of the vessels across the intersection while being penalized for not yielding to the agent coming from the right-hand side.

\subsubsection{Ancillary Controllers}\label{subsec::AC}
To help switch out of local minima and improve sampling efficiency, four ancillary controllers are sampled using the proposed Biased-MPPI.

\paragraph{Go-Slow} a sequence of inputs commanding a small amount of thrust to the vessel's side thrusters.

\paragraph{Go-Fast} commands a large thrust.

\paragraph{Braking} gives a zero velocity reference.

\paragraph{Go-to-Goal} computes a velocity reference that takes each vessel towards its corresponding local goal at each time step of the planning horizon.

The velocity references proposed by the \textit{Braking} and \textit{Go-to-Goal} maneuvers are converted to input thrusts with a linear $\mathcal{H}_\infty$ controller, which is robust to model non-linearities, designed using the \verb"musyn" command in Matlab.

\subsubsection{Results}

With an initial velocity of zero, each agent anticipates an unobstructed intersection crossing. This expectation is based on a constant velocity prediction, as they assume the opposing agent will remain stationary.
In~\Cref{fig:crossing_vanilla}, the classic IA-MPPI fails to switch from planning to cross first to a slower maneuver that yields since all of the samples are taken around the previous plan, leading to a collision.
In~\Cref{fig:crossing_biased}, our Biased-IA-MPPI approach can swiftly switch between modes when it becomes evident that the vessel with the right-of-way will cross the intersection.

In~\Cref{tab::tight_crossing}, we see that in 50 experiments, our Biased-IA-MPPI achieves zero collisions and rule violations for any number of samples, compared to the IA-MPPI based on the classical MPPI sampling scheme, which results in several.
Thanks to the ancillary controllers, our Biased-IA-MPPI also travels straight to the goal, reducing the distance traveled.
While our Biased-IA-MPPI has a lower variance in arrival times, it is not always faster on average. This confirms the results proved in~\cref{eq::bias}, i.e. the \textit{Braking} and \textit{Go-Slow} maneuvers are biasing towards a slower trajectory.

\begin{table}[ht!]
\begin{center}
\caption{Results of 50 crossings for an increasing number of samples K. Metrics are reported for successful runs.} \label{tab::tight_crossing}
\setlength{\tabcolsep}{1.03pt}
\begin{tabular}{c l c c c c c}
\toprule
 \multirow{3}{*}{\textbf{K}} & \multirow{3}{*}{\textbf{Method}} & \multirow{3}{*}{\textbf{Collisions}}  &
 \textbf{Experiments} & \textbf{Average} & \textbf{Average} \\
 &                                  &                                       &
 \textbf{With Rule}   & \textbf{Time to} & \textbf{Distance} \\
 &                                  &                                       &
 \textbf{Violations}  & \textbf{Arrival [s]} & \textbf{Traveled [m]} \\

\midrule
\parbox[t]{2mm}{\multirow{2}{*}{\rotatebox[origin=c]{90}{50}}} 
 & IA-MPPI &  4 & 9  & \textbf{16.41} $\pm$ 10.10                           & 21.89 $\pm$ 8.433\\
 & Biased-IA-MPPI &     \textbf{0} & \textbf{0}  & 17.64 $\pm$ \textbf{3.128}         & \textbf{19.13} $\pm$ \textbf{2.466}\\
\midrule
\parbox[t]{2mm}{\multirow{2}{*}{\rotatebox[origin=c]{90}{200}}} 
 & IA-MPPI &  10 & 4 & 12.77 $\pm$ 9.323                                    & 19.99 $\pm$ 10.16\\
 & Biased-IA-MPPI &     \textbf{0} & \textbf{0}  & \textbf{12.66} $\pm$ \textbf{1.902}& \textbf{18.07} $\pm$ \textbf{2.012}\\
\midrule
\parbox[t]{2mm}{\multirow{2}{*}{\rotatebox[origin=c]{90}{500}}}
 & IA-MPPI &  7 & 11 & \textbf{11.02} $\pm$ 2.731                           & 18.70 $\pm$ 3.518\\
 & Biased-IA-MPPI &     \textbf{0} & \textbf{0}  & 11.43 $\pm$ \textbf{1.541}         & \textbf{17.58} $\pm$ \textbf{1.880}\\
\midrule
\parbox[t]{2mm}{\multirow{2}{*}{\rotatebox[origin=c]{90}{1000}}}
 & IA-MPPI &  10 & 15 & 11.78 $\pm$ 3.823                                   & 19.31 $\pm$ 3.539\\
 & Biased-IA-MPPI &     \textbf{0} & \textbf{0}  & \textbf{11.00} $\pm$ \textbf{1.309}& \textbf{17.35} $\pm$ \textbf{1.625}\\
\midrule
\parbox[t]{2mm}{\multirow{2}{*}{\rotatebox[origin=c]{90}{2000}}}
 & IA-MPPI &  7 & 15 & 11.10 $\pm$ 4.101                                    & 19.72 $\pm$ 5.038\\
 & Biased-IA-MPPI &     \textbf{0} & \textbf{0}  & \textbf{10.68} $\pm$ \textbf{1.245}& \textbf{17.27} $\pm$ \textbf{1.716}\\
\bottomrule 
\end{tabular}   
\end{center}
\end{table}

\subsection{Interaction-Aware Planning with Four Vessels}
\begin{figure}[t]
    \centering
    \includegraphics[width=0.2\textwidth]{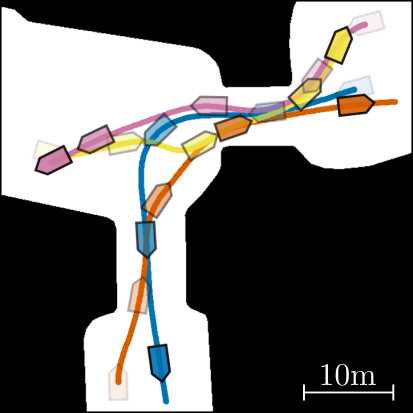}
    \caption{Four agents navigating in the Herengracht. \href{https://autonomousrobots.nl/paper_websites/biased-mppi}{Video}.}
    \label{fig:herengracht}
\end{figure}

To further test Biased-IA-MPPI, we run 50 experiments with randomized initial conditions and goals, where four agents have to navigate in cooperation in the Herengracht, an urban canal in Amsterdam, challenging due to its narrow sections under two bridges. The canal map and an example of successful navigation are shown in~\Cref{fig:herengracht}.

\subsubsection{Ancillary Controllers}

We use all of the ancillary controllers described in~\Cref{subsec::AC}.
Additionally, we use a learning-based trajectory prediction model adapted and trained for urban vessels~\cite{jansma_interaction-aware_2023_damn_alvaro}. However, we do not use this model for predictions. We track the trajectories it provides with an $\mathcal{H}_\infty$ controller to generate input sequences, which Biased-MPPI can consider in its sampling scheme.

\subsubsection{Results}
In~\Cref{tab::results}, results from 50 experiments show that with 50 samples, our Biased-IA-MPPI is cautious, leading to 10 deadlocks, possibly biased by the \textit{Braking} maneuver. In contrast, the conventional IA-MPPI approach, without the ancillary controller, results in 16 collisions.

\begin{table}[h!]
\begin{center}
\caption{Results for 50 runs of four-agent experiments in the Herengracht with randomized initial conditions and goals for an increasing number of samples K.} \label{tab::results}
\setlength{\tabcolsep}{3pt}
\begin{tabular}{c l c c c c c}
\toprule
 \multirow{3}{*}{\textbf{K}} & \multirow{3}{*}{\textbf{Method}} & \multirow{3}{*}{\textbf{Successes}} & \multirow{3}{*}{\textbf{Deadlocks}} & \multirow{3}{*}{\textbf{Collisions}} & \textbf{Experiments} \\
 &                                  &  &  &  & \textbf{With Rule} \\
 &                                  &  &  &  & \textbf{Violations} \\

\midrule
\parbox[t]{2mm}{\multirow{2}{*}{\rotatebox[origin=c]{90}{50}}} 
 & IA-MPPI &  34   &     0    &    16     &     22 \\
 & Biased-IA-MPPI &     \textbf{40}   &    10    &     \textbf{0}     &     \textbf{18} \\
\midrule
\parbox[t]{2mm}{\multirow{2}{*}{\rotatebox[origin=c]{90}{200}}} 
 & IA-MPPI &  43    &    1    &     6    &     34  \\
 & Biased-IA-MPPI &     \textbf{46}    &    1    &     \textbf{3}    &     \textbf{28}  \\
\midrule
\parbox[t]{2mm}{\multirow{2}{*}{\rotatebox[origin=c]{90}{500}}}
 & IA-MPPI &  47   &     0    &     3    &     36  \\
 & Biased-IA-MPPI &     \textbf{49}    &    0    &     \textbf{1}    &     \textbf{35}  \\
\midrule
\parbox[t]{2mm}{\multirow{2}{*}{\rotatebox[origin=c]{90}{1000}}}
 & IA-MPPI &  45    &    0    &     5    &     36  \\
 & Biased-IA-MPPI &     \textbf{50}    &    0    &     \textbf{0}    &     \textbf{33}  \\
\midrule
\parbox[t]{2mm}{\multirow{2}{*}{\rotatebox[origin=c]{90}{2000}}}
 & IA-MPPI &  47    &    0    &     3    &     36  \\
 & Biased-IA-MPPI &     \textbf{49}    &    0     &    \textbf{1}    &     \textbf{34}  \\
\bottomrule 
\end{tabular}   
\end{center}
\end{table}

As the number of samples increases, the bias from the ancillary controllers diminishes, causing Biased-IA-MPPI to behave less conservatively. Consequently, the number of deadlocks approaches zero, but a few collisions may occur.
With both methods, over half of the successful experiments incur at least a rule violation. In these crowded scenes, violations are common, e.g., not stopping to yield to an agent with priority when it is still relatively far away.
Still, in both collision counts and the number of experiments resulting in rule violations, our Biased-IA-MPPI consistently outperforms IA-MPPI using the traditional sampling method.

\begin{figure}[t]
    \centering
    \includegraphics[width=0.48\textwidth]{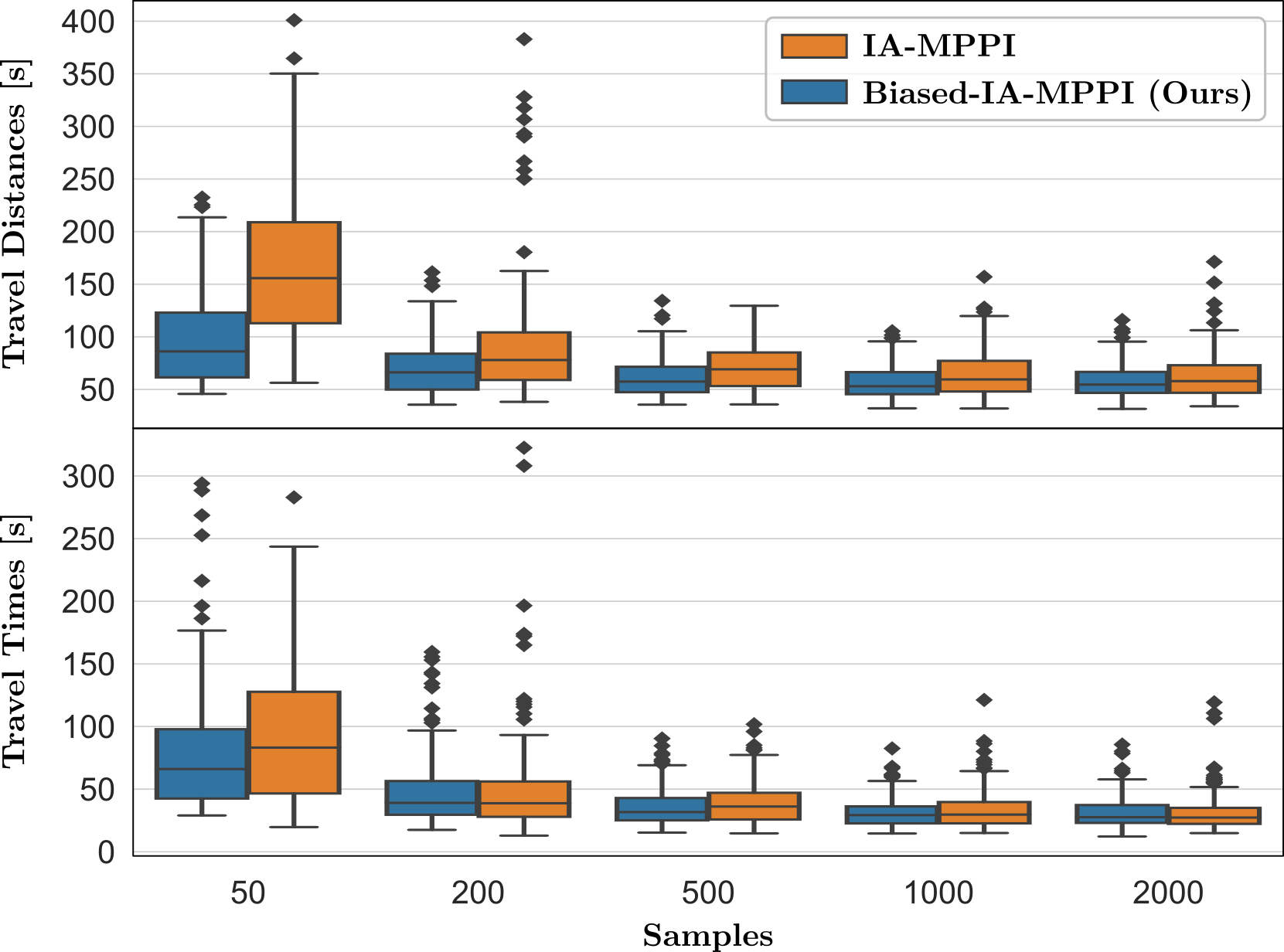}
    \caption{Agents' traveled distance and travel time over 50 experiments in the Herengracht. Metrics are reported for experiments that were successful with both methods.}
    \label{fig:Travel_distance}
\end{figure}

\Cref{fig:Travel_distance} displays both methods' quartiles, min, max, and outliers of successful experiments. The ancillary controllers direct the sampling distribution towards lower-cost areas of the state space, significantly reducing travel distances. Despite this, as predicted by~\cref{eq::bias}, Biased-IA-MPPI also exhibits a bias towards slightly slower movement due to ``Braking" and ``Go-Slow" maneuvers, resulting in similar travel times as the regular IA-MPPI.

\section{Real-World Motion Planning Experiment}
A Clearpath Jackal robot attempts to drive to a goal as fast as possible ($\sim 2 m/s$) while avoiding a box. Halfway through, the box is thrown in front of the robot. The position and the velocity of the box and the robot are estimated using information from a motion capture system. The velocity-controlled robot is modeled as a unicycle, and the box's position is propagated through the planning horizon using a constant velocity model. The cost function is defined as,
\begin{equation}
    C_j(x(t)) = ||p_{t,r} - p_g || + 100 (||p_{t,r} - p_{t,b}|| < 0.5)
\end{equation}
where $p_{t,r}$, $p_g$ and $p_{t,b}$ are the position of the robot, the goal, and the box, respectively, at time $t$. Controllers run at $10Hz$, plan $T_H=50$ steps ahead ($5s$), with $K=300$ samples, covariance $\Sigma_s=0.5 \cdot I_{2\times2}$, $\eta_{min}=5$ and $\eta_{max}=10$.

\begin{figure}[t]
    \medskip
    \centering
    \includegraphics[width=0.45\textwidth]{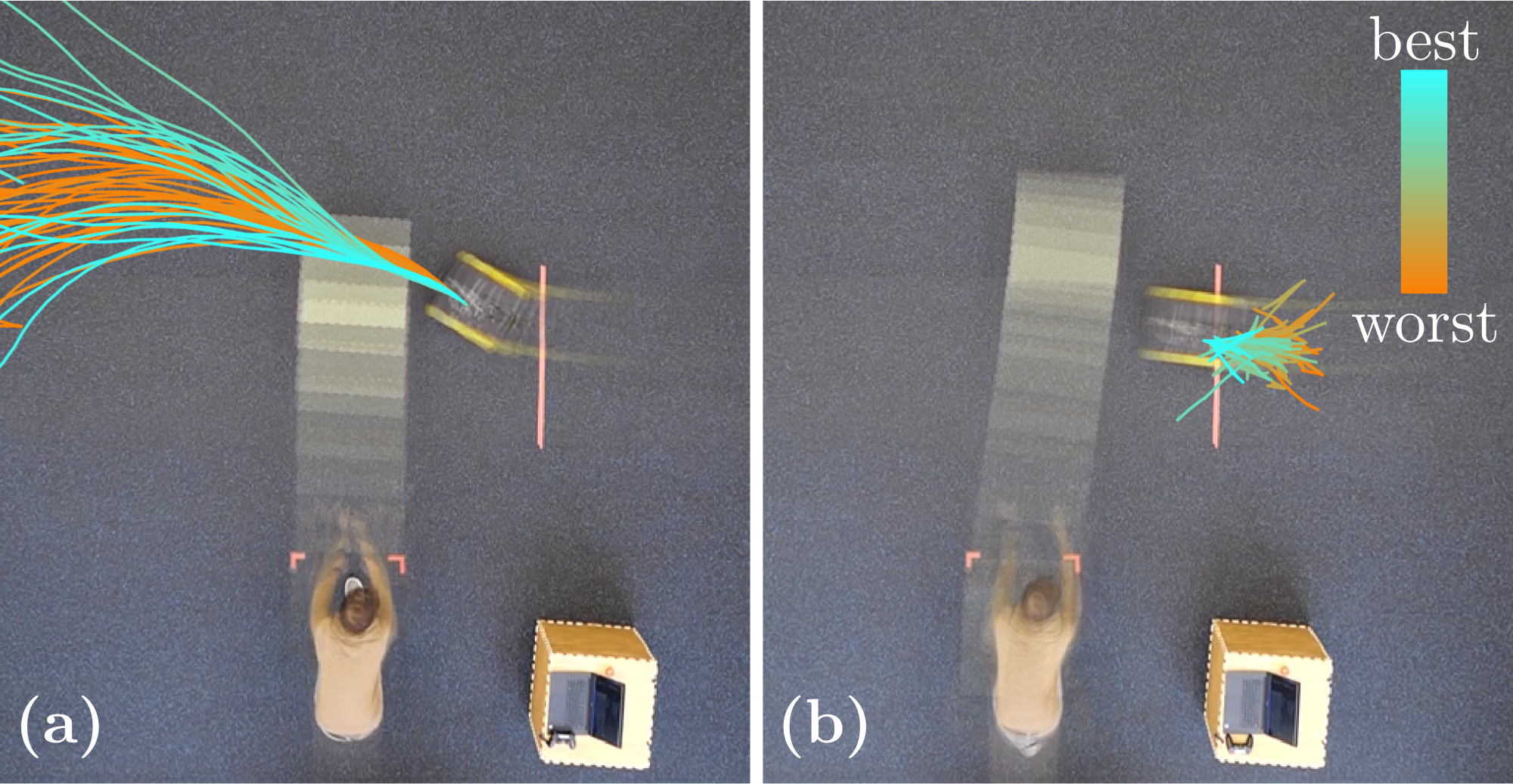}
    \caption{Visualized are the top 50 sampled trajectories, color-graded by their cost. (a) Classic MPPI is about to crash. (b) Our Biased-MPPI avoids collision. See \href{https://autonomousrobots.nl/paper_websites/biased-mppi}{video}.}
    \label{fig:real_world}
\end{figure}

\subsubsection{Ancillary Controllers}
We sample a \textit{Braking} maneuver, i.e., a zero velocity reference throughout the horizon.

\subsubsection{Results}
\Cref{fig:real_world} shows the top 50 sampled trajectories sampled by (a), MPPI, and (b), our proposed Biased-MPPI. When the box is unexpectedly thrown in front of the robot, MPPI only samples trajectories that collide with the box. Given the cost function, MPPI prefers the samples that remain in collision for the least time. On the other hand, sampling also a zero velocity reference, Biased-MPPI quickly converges to a braking maneuver, avoiding the collision altogether. MPPI resulted in six collisions in over ten experiments, while Biased-MPPI resulted in none.


\section{Conclusions}\label{sec::conclusion}
\label{sec:conclusions}
In this paper, we have derived a sampling scheme for Model Predictive Path Integral (MPPI) control that removes computationally problematic terms and allows for the design of arbitrary sampling distributions as long as a bias in the solution is allowed.
We proposed using classical and learning-based ancillary controllers for several control and motion planning experiments to bias the sampling distribution and achieve more efficient sampling and better performances.
We demonstrated how the proposed algorithm can act as a control fusion scheme, taking suggestions from an arbitrary number of controllers and improving upon them.
The resulting Biased-MPPI was shown to be better performing and more robust to model uncertainties compared to classical controllers and the baseline MPPI method, achieving faster swing-ups for a rotational inverted pendulum as well as safer, closer to optimal trajectories in interaction-aware motion planning experiments in constrained multi-agent environments, all while requiring less samples.
The overall gains in safety, performance, and sample efficiency come at the expense of a potentially harmful bias, as shown with the sampling of \textit{Braking} and \textit{Go-Slow} maneuvers, which can result in slower trajectories.
In the future, our approach could be employed as a potential solution to complex multi-modal problems. For example, a higher-level task planner could propose several ancillary controllers and alternative plans, which could all be sampled to achieve global solutions.


\bibliographystyle{IEEEtran}
\bibliography{IEEEabrv,mybib,references}

\end{document}